\documentclass[conference]{IEEEtran} % or journal, or peerreview, etc.

\usepackage{graphicx}
\usepackage{amsmath,amssymb}
\usepackage{cite}
\usepackage{algorithm}
\usepackage{algorithmic}
\usepackage{adjustbox}
\usepackage{multirow}
\usepackage[table]{xcolor}
\usepackage{float}
\usepackage{subcaption}
\usepackage{lscape}
\usepackage{balance}
\usepackage{url}
\usepackage{booktabs}
\usepackage{hyperref}

\title{DepMicroDiff: Diffusion-Based Dependency-Aware Multimodal Imputation for Microbiome Data}

\author{
    \IEEEauthorblockN{Rabeya Tus Sadia}
    \IEEEauthorblockA{
        Department of Computer Science \\
        University of Kentucky \\
        rabeya.sadia@uky.edu
    }
    \and
    \IEEEauthorblockN{Qiang Cheng\footnotemark *}
    \IEEEauthorblockA{
        Department of Computer Science, \\
        Institute for Biomedical Informatics \\
        University of Kentucky \\
        qiang.cheng@uky.edu
    }
}

\begin{document}

\maketitle
\renewcommand{\thefootnote}{\fnsymbol{footnote}}
\footnotetext{* Corresponding author: Qiang Cheng}
\renewcommand{\thefootnote}{\arabic{footnote}}
% -------------------- Abstract --------------------
\begin{abstract}
Microbiome data analysis is essential for understanding host health and disease, yet its inherent sparsity and noise pose major challenges for accurate imputation, hindering downstream tasks such as biomarker discovery. Existing imputation methods, including recent diffusion-based models, often fail to capture the complex interdependencies between microbial taxa and overlook contextual metadata that can inform imputation. We introduce DepMicroDiff, a novel framework that combines diffusion-based generative modeling with a Dependency-Aware Transformer (DAT) to explicitly capture both mutual pairwise dependencies and autoregressive relationships. DepMicroDiff is further enhanced by VAE-based pretraining across diverse cancer datasets and conditioning on patient metadata encoded via a large language model (LLM). Experiments on TCGA microbiome datasets show that DepMicroDiff substantially outperforms state-of-the-art baselines, achieving higher Pearson correlation (up to 0.712), cosine similarity (up to 0.812), and lower RMSE and MAE across multiple cancer types, demonstrating its robustness and generalizability for microbiome imputation.
\end{abstract}

% -------------------- Keywords --------------------

% -------------------- Sections --------------------
\section{Introduction}
% [Insert Introduction Text Here]
Microbiome data analysis plays a critical role in understanding host health, disease progression, and therapeutic response, particularly in contexts such as cancer progression, gut-brain interactions, and immunotherapy \cite{dohlman2021cancer}. However, microbiome datasets, derived from 16S rRNA or metagenomic sequencing, are notoriously sparse and noisy due to limitations in sequencing technologies, biological variability, and compositional constraints. Missing or zero-valued entries can severely compromise downstream tasks such as clustering, classification, and biomarker discovery. Effective imputation of such data remains a major and persistent challenge.

Traditional imputation techniques, including K-nearest neighbors (KNN) and matrix factorization, often fail to capture the high-dimensional, nonlinear, and context-specific structure of microbiome profiles. In response, deep learning models, such as autoencoders, variational inference frameworks, and transformer-based architectures, have been increasingly adopted in omics research. Methods like DeepImpute \cite{deepimpute}, DCA \cite{dca}, and scVI \cite{scvi} have shown strong performance on gene expression or methylation data, and recent efforts have adapted these models for microbiome applications \cite{shi2024pretrained}. Nonetheless, these approaches often underutilize crucial inductive biases, such as complex interdependencies among microbial taxa (reflecting ecological interactions), and auxiliary biological metadata (e.g., tissue type or disease stage), which can provide essential context for accurate imputation.

Meanwhile, diffusion models have emerged as powerful generative frameworks across vision, language, and bioinformatics, capable of modeling complex data distributions via iterative denoising \cite{shi2024pretrained, zhong2024synthesizing, senane2024self}. Despite their success in genomics and single-cell RNA-seq, their potential for microbiome data, characterized by extreme sparsity and structured interdependencies, remains largely unexplored. Modeling both long-range and localized dependencies in a scalable manner is critical for effective and interpretable imputation in this setting.

To address these challenges, we propose \textit{DepMicroDiff}, a novel imputation framework that integrates diffusion-based generative modeling with dependency-aware and autoregressive mechanisms tailored for microbiome data. DepMicroDiff introduces a \textit{Dependency-Aware Transformer (DAT)} module to explicitly capture biological and sequential dependencies. To improve generalization in low-data regimes, we incorporate a \textit{variational autoencoder (VAE)-based pretraining} strategy that learns structured latent representations across diverse tissue types. Moreover, DepMicroDiff conditions on patient-level metadata (e.g., tissue or cancer type) using a generalized large language model (LLM) encoder, enhancing contextual awareness at the sample level.

In brief, our contributions are summarized as follows:
\begin{itemize}
\vspace{-5pt}
    \item We introduce DepMicroDiff, a novel diffusion-based imputation framework equipped with a Dependency-Aware Transformer (DAT) to model both autoregressive and mutual dependencies, enabling the capture of long-range feature interactions in sparse microbiome datasets.
    \item We develop a VAE-based pretraining scheme to improve cross-tissue generalization and reduce overfitting in low-sample regimes.
    \item We condition imputation on auxiliary patient metadata using a generalized LLM encoder, enhancing instance-level semantic awareness.
    \item Through extensive experiments on multiple cancer-associated microbiome datasets, we demonstrate that DepMicroDiff outperforms state-of-the-art baselines under both supervised and zero-shot evaluation settings.
\end{itemize}
To contextualize our contributions, we will review existing approaches to microbiome data imputation below, highlighting their limitations in capturing complex microbial dependencies.
\section{Related Work}
% [Insert Related Work Text Here]
Missing value imputation is a crucial preprocessing step in microbiome analysis, as sparsity often arises due to detection limits, sequencing depth, or biological variability in datasets like 16S rRNA or metagenomic profiles. Traditional methods such as K-Nearest Neighbors (KNN) have been widely used for their simplicity but fail to capture the nonlinear and high-dimensional structure of microbiome data, often producing oversimplified imputations that ignore ecological interactions between microbial taxa.

To overcome these limitations, deep learning-based methods have been introduced in omics data analysis, initially for single-cell transcriptomics and gradually adapted to microbiome contexts \cite{shi2024pretrained}. DeepImpute \cite{deepimpute} employs a deep neural network that models gene–gene dependencies in a subset-wise manner, improving scalability but potentially overlooking long-range dependencies critical for microbial co-occurrence patterns. Similarly, AutoImpute \cite{autoimpute} uses autoencoders to learn nonlinear embeddings for reconstructing missing expression values, but it lacks explicit modeling of causal relationships. DCA \cite{dca} introduces a deep count autoencoder tailored for overdispersed count data, achieving strong denoising performance in single-cell RNA-seq but requiring adaptation for microbiome-specific compositional constraints.

Generative approaches further enhance imputation capabilities. CpG Transformer \cite{cpg}, originally developed for single-cell methylome data, demonstrates that attention mechanisms can effectively model structured patterns of missingness. However, its applicability to microbiome data is limited by the absence of microbial-specific priors. scVI \cite{scvi}, a variational autoencoder (VAE) framework, incorporates probabilistic priors to model transcriptomic variation and has become popular for imputation and clustering in scRNA-seq. Although these models exploit the high-dimensional and sparse characteristics shared with microbiome data \cite{shi2024pretrained}, their direct application necessitates addressing domain-specific challenges such as zero-inflation and ecological dependencies.

In the microbiome domain, deep generative models are emerging as promising alternatives. DeepMicroGen \cite{deepmicrogen} employs a GAN-based framework with recurrent networks to impute longitudinal microbiome data, capturing temporal dynamics but suffering from potential training instability. More recently, mbVDiT \cite{shi2024pretrained} combines a pretrained transformer backbone with a conditional diffusion model guided by observed microbiome profiles and patient metadata, achieving competitive performance in context-aware imputation. However, it does not explicitly model causal relationships or co-occurrence patterns among microbial taxa, which are critical for capturing ecological and functional dependencies \cite{bucci2016microbial}.

Despite these advances, existing methods rarely capture the complex interdependencies and co-occurrence  patterns among microbial taxa, which are essential for understanding ecological and functional relationships in microbiome data. Our proposed DepMicroDiff addresses this gap by integrating diffusion-based generative modeling with dependency-aware mechanisms, as detailed in the following sections.
\section{Dependencies in Microbiome Data}
\begin{figure}[H]
    \centering
    \includegraphics[width=0.48\textwidth, height=0.15\textwidth]{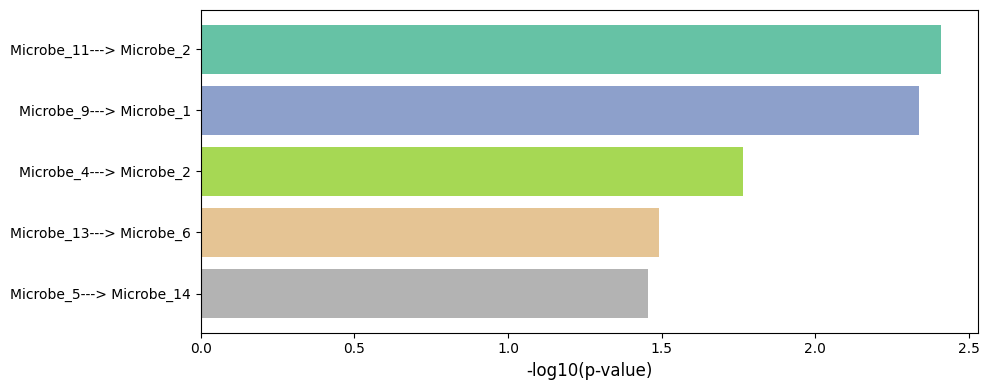}
    \caption{Top Granger causal relationships among high-variance microbes in the COAD dataset. Bar lengths indicate $-\log_{10}$-transformed p-values, reflecting the strength of statistical dependency. For example, Microbe\_11 $\rightarrow$ Microbe\_2 denotes that temporal variations in Microbe\_11 significantly predict changes in Microbe\_2.}
    \label{fig:causal}
    \vspace{-11pt}
\end{figure}

% [Insert Analysis and Figures]
We propose to leverage dependency relationships among microbial taxa to enhance microbiome data modeling using deep neural networks. To this end, we conducted a pairwise Granger causality analysis to uncover the top 5 most variable pairs of microbes with significant dependencies from the microbiome abundance matrix. Granger causality is a time-series-based statistical framework that assesses whether the historical values of one variable can improve the prediction of another, thereby inferring potential directional influence. Figure~\ref{fig:causal} illustrates the top five causal pairs of  Colon Adenocarcinoma
(COAD) dataset with $p$-values below 0.05, ranked by $-\log_{10}(p\text{-value})$ to emphasize statistical strength. Each horizontal bar represents a directed edge from a causal microbe to an effect microbe. Notably, we observed that Microbe\_11 significantly predicts Microbe\_2, and both Microbe\_9 and Microbe\_4 also influence downstream targets, suggesting hierarchical or regulatory patterns in the microbial community. These results highlight non-random, directional structure in microbial dynamics, motivating the development of our proposed conditional diffusion framework, which explicitly incorporates dependency relationships and patient metadata to enhance microbiome data imputation and interpretation.

\begin{figure}[H]
    \centering
    \includegraphics[width=0.5\textwidth]{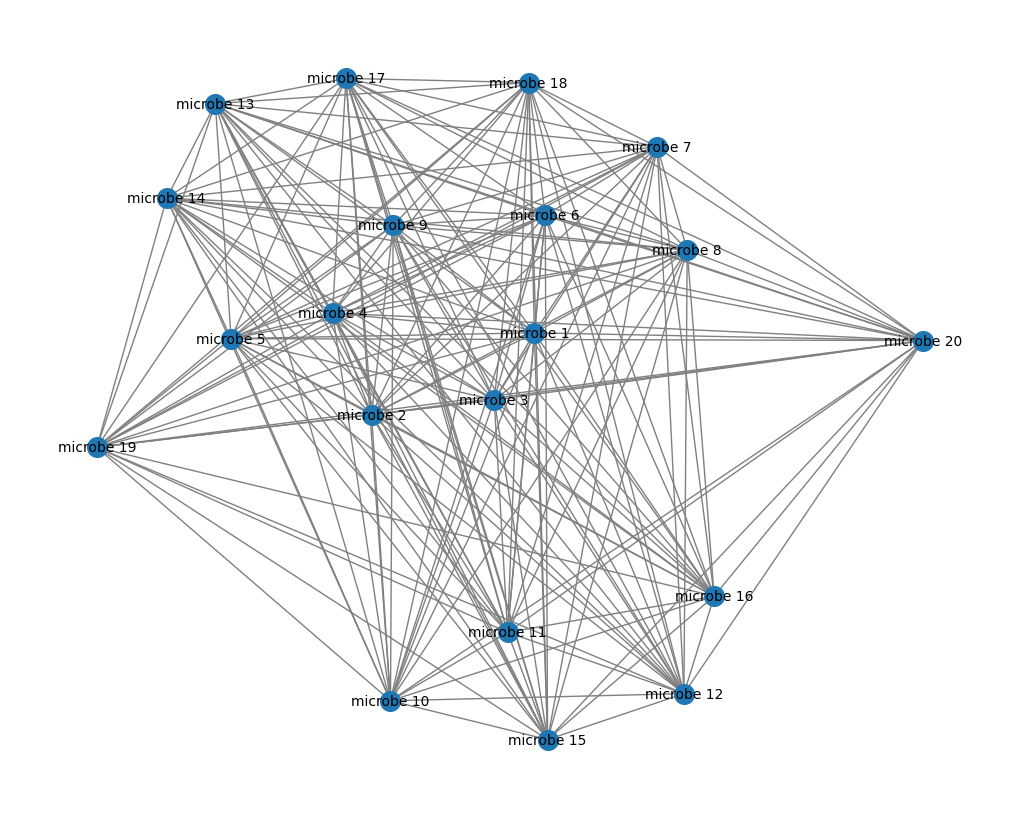}
    \caption{Microbial dependency network based on mutual information. Each node represents a microbe, and edges represent pairwise mutual information above a predefined threshold, indicating potential statistical dependency between microbial abundances.}
    \label{fig:microbe_network}
\end{figure}

While Granger analysis revealed only a limited number of significant directional dependencies (5 pairs with p < 0.05), this sparsity motivated us to explore more general dependency structures. We therefore employed mutual information, an information-theoretic metric, to quantify the dependencies between each pair of microbes. 
Figure~\ref{fig:microbe_network} illustrates the mutual information-based dependency network of COAD dataset among the top 20 most variable microbes. Nodes represent individual microbes, and edges connect microbe pairs exhibiting statistically significant dependencies based on mutual information analysis (p < 0.05, permutation test). The dense connectivity pattern reveals the intricate interdependencies among microbes, potentially reflecting co-occurrence or interaction structures. This graph structure motivates the exploration of dependency modeling approaches in our generative frameworks.
\section{Overview of Our Model}
\begin{figure*}[htp]
\centering
\includegraphics[width=0.9\textwidth]{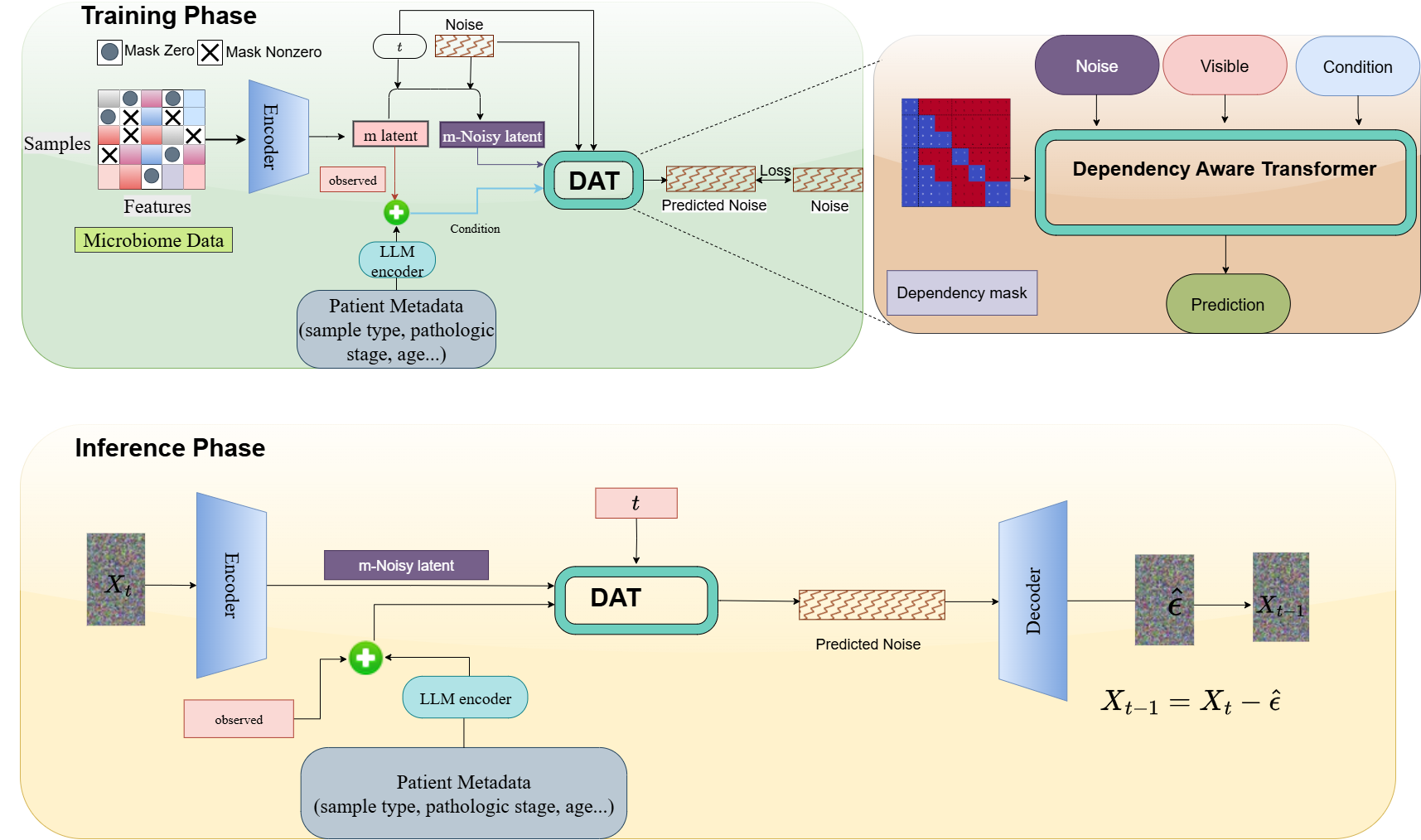}
\caption{
Overview of the DepMicroDiff architecture for microbiome data imputation using cross-modality conditioning and diffusion modeling.
\textbf{Training Phase:} Partially masked microbiome data is encoded into a latent space utilizing a pretrained VAE encoder with the Diffusion model, including autoregressive components. The denoising model (\textsc{DAT}: Dependency-Aware Transformer) conditions on LLM-encoded patient metadata and the observed portion of the latent to predict the noise. A dependency-guided mask enforces attention to statistically dependent features during training. 
\textbf{Inference Phase:} Starting from Gaussian noise, the model iteratively denoises the latent representation using the same conditioning signals and reconstructs the full microbiome profile via the decoder.
}
\label{fig:method}
\end{figure*}

% [Describe VAE, Conditioning, Diffusion, DAT, etc. Include Figures]
We present a detailed overview of the components within our proposed DepMicroDiff model. The architecture is composed of the following key modules:

\paragraph{\bf Latent space representation}
To obtain latent space representations, we employ a VAE pretrained encoder architecture, as illustrated in Figure~\ref{fig:method}, the encoder $E$ maps the masked microbiome input $x_{mb}^i \in \mathbb{R}^{p \times q}$, where $p$ represents number of samples and $q$ represents microbe features, into a latent representation $\hat{x}_{mb}^i \in \mathbb{R}^{d}$, preserving the inherent variation in the data, where $d$ denotes the dimensionality of the latent space. The encoder $E$ and corresponding decoder are pretrained in a Variational AutoEncoder (VAE) framework. This process can be formally defined as:
\begin{equation}
    \hat{x}_{mb}^i = E(x_{mb}^i),
\end{equation}
The resulting embedding, referred to as the m-latent in Figure~\ref{fig:method}, serves as the initial input to the diffusion process. 
\paragraph{\bf Conditioning Mechanism}
To guide the imputation process, our model leverages both the partially observed microbiome data and rich patient metadata (e.g., sample type, pathologic stage, age) as conditioning information. Let \( m_0 \) denote the initial latent representation of the microbiome data, obtained from the pretrained VAE encoder. We define the observed portion of the latent as:
\begin{equation}
m_0^{c} = (E_{n,d} - m) \odot m_0,
\end{equation}
where \( m \in \{0,1\}^{n \times d} \) is a binary mask indicating masked (1) and observed (0) entries, and \( E_{n,d} \) is an all-ones matrix of the same shape as \( m_0 \). The symbol \( \odot \) denotes element-wise multiplication. This observed latent \( m_0^{c} \) is used as part of the conditioning signal. Additionally, the patient metadata is tokenized and encoded via a pretrained large language model (LLM). For our study, we used the Bidirectional Encoder Representations from Transformers (BERT) as the large language model encoder. However, other LLM-based encoders such as BioBERT, ClinicalBERT, or PubMedBERT can also be employed for text-based conditioning. The resulting embedding is projected and fused with \( m_0^{c} \), forming a comprehensive conditioning vector that is supplied to the diffusion model throughout training and inference. This design enables our model to incorporate both structural and biological priors during denoising.
\paragraph{\bf Latent Noise Modeling for Diffusion}
We construct the noisy input referred to as the m-Noisy latent in Figure~\ref{fig:method} by applying a forward diffusion process to the clean latent representation $\hat{x}_{mb}^i$. This process involves gradually injecting random Gaussian noise into $\hat{x}_{mb}^i$ over a sequence of steps, forming a Markov chain. At each diffusion step $t$, which also represents the noise level, the noisy latent variable $\hat{x}_{mb}^{i,t}$ is conditionally dependent on its previous state $\hat{x}_{mb}^{i,t-1}$ through the following transition distribution:
\begin{equation}
\label{eq-transition}
    q(\hat{x}_{mb}^{i,t} | \hat{x}_{mb}^{i,t-1}) = \mathcal{N}(\hat{x}_{mb}^{i,t}; \sqrt{1 - \beta_t} \, \hat{x}_{mb}^{i,t-1}, \beta_t \mathbf{I}),
\end{equation}
where $\beta_t$ denotes the cosine variance schedule at time step $t$, and $\mathbf{I}$ is the identity matrix. The sequence $\{\beta_t\}_{t=1}^T$ is monotonically increasing, i.e., $\beta_1 < \beta_2 < \cdots < \beta_T$, ensuring that noise is introduced progressively. The full forward diffusion process is defined by chaining these transitions, resulting in the joint distribution:
\begin{equation} \label{eq:diffusion-factorization}
    q(\hat{x}_{mb}^{i,0:T}) = q(\hat{x}_{mb}^{i,0}) \prod_{t=1}^T q(\hat{x}_{mb}^{i,t} | \hat{x}_{mb}^{i,t-1}).
\end{equation}
\paragraph{\bf{Variational Autoencoder Pretraining}}
The Variational AutoEncoder (VAE) module comprises an encoder and a decoder, following the conventional architecture and functionality of standard VAEs. Its primary role in our framework is to learn the latent distributions and feature representations from datasets corresponding to various cancer types.
To this end, we first pre-train the VAE using data from four distinct cancer types. The learned parameters are then transferred to initialize the VAE component within the DepMicroDiff model. Subsequently, DepMicroDiff is trained on a target dataset that is excluded from the initial pre-training datasets, ensuring no data leakage during fine-tuning.
Consider, we are given an input data $\mathit{X} \in \mathbb{R}^{n \times d}$, representing a set of $n$ samples with $d$ features, the encoder maps the input to a latent distribution parameterized by the mean $\mu_i$ and variance $\sigma_i^2$. The latent representation \( z \) is derived through the reparameterization trick, expressed as:

\begin{equation}
    z = \mu + \sigma \cdot \epsilon, \quad \text{where} \quad \sigma = e^{1/2 \log \sigma^2}, \quad \epsilon \sim \mathcal{N}(0, I).
\end{equation}

The decoder utilizes this latent variable to reconstruct the input data, aiming to retain the core structure and patterns of the original features. Despite handling two distinct input modalities, we employ a single shared decoder, promoting parameter efficiency within the architecture.

The VAE's loss function is composed of two parts: a reconstruction loss and a Kullback–Leibler (KL) divergence term. The reconstruction loss is defined as:

\begin{equation}
    \mathcal{L}_{\text{recon}} = \| X - \hat{X} \|^2,
\end{equation}

\noindent
which ensures the output is a faithful reconstruction of the input. The KL divergence term,

\begin{equation}
    \mathcal{L}_{\text{KL}} = -\frac{1}{2} \sum \left( 1 + \log \sigma^2 - \mu^2 - \sigma^2 \right),
\end{equation}

\noindent
encourages the learned latent variables to follow a standard normal distribution, thereby regularizing the latent space.

The final loss combines both components as follows:

\begin{equation}
    \mathcal{L} = \mathcal{L}_{\text{recon}} + \mathcal{L}_{\text{KL}},
\end{equation}

\noindent
striking a balance between reconstruction fidelity and latent distribution regularization. This VAE pretraining mechanism facilitates effective representation learning with structured latent embeddings.
\paragraph{\bf{Diffusion Sampling Strategy}} 
Our model employs a carefully designed diffusion sampling strategy that balances computational efficiency with predictive accuracy. The forward diffusion process spans \( T \) timesteps (default \( T = 1000 \)), wherein Gaussian noise is progressively added at each step \( t \) following a predefined variance schedule \( \beta_t \). Rather than using a uniform timestep across all autoregressive (AR) steps, we introduce a more flexible and adaptive sampling approach:

1. \textbf{Base Strategy:} For each autoregressive step \( s \), we sample from a diverse set of diffusion timesteps to capture varying noise scales. This allows the model to learn dependencies that manifest across a spectrum of regulatory strengths and complexities.

2. \textbf{Efficient Sampling:} To minimize computational cost while preserving performance, we incorporate three sampling modes: \textit{Full sampling} (\(T\)) utilizes the complete diffusion sequence; \textit{Fractional sampling} (\(1/nT\)) selects evenly spaced timesteps from the full set, where \(n \in \{2, 3, 4, 20\}\); and \textit{Adaptive sampling} dynamically modifies the sampling frequency based on the relevance of each AR step.

The sampling process is aligned with the AR framework via the transition distribution (Equation~\ref{eq-transition}). Each AR step adopts its specific sampling schedule while reusing clean tokens generated in prior steps. 
\paragraph{\bf Mixing Diffusion and Autoregression}  
\begin{figure}[t]
    \centering
    \includegraphics[width=0.5\linewidth]{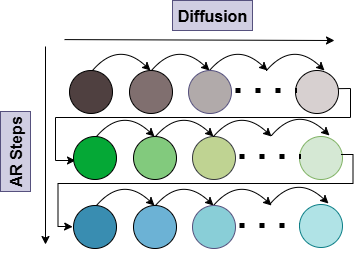} % Replace with your image path 
    \caption{Diffusion with Autoregression. Tokens are simultaneously denoised while maintaining dependency relations from previous tokens.}
    \label{fig:cat_diff}
    \vspace{-12pt}
\end{figure}

Although diffusion models are powerful for data generation, they often lack the ability to model sequential dependencies effectively, an area where autoregressive (AR) models excel. To overcome this limitation, we propose the Dependency-Aware Transformer (DAT) module, which fuses the strengths of both approaches to more faithfully represent microbial associations. The DAT module processes microbial features in a sequential autoregressive manner, emulating hierarchical dependencies and regulatory cascades commonly observed in microbial community dynamics and host-microbe interactions. This autoregressive ordering is governed by a dependency-guided attention mechanism that enforces directional flow and dependency relationship, allowing earlier features to influence subsequent ones, mirroring real-world biological regulation.
While the DAT architecture draws inspiration from image-based methods such as~\cite{deng2024causal}, our formulation is substantially adapted to handle microbiome data. DAT addresses these challenges via four core components: (1) a modified, conditional, dependency-aware attention mechanism tailored to microbial data rather than categorical image labels; (2) tokenization of microbe-level compressed features, where each token corresponds to a microbe rather than a patch; (3) a VAE-based pretraining scheme using diverse cancer-type microbiome datasets to support generalizable latent diffusion; and (4) a conditioning strategy that combines partially observed latent features with BERT LLM-encoded patient metadata for biologically informed denoising.

To enhance robustness and flexibility, the AR step sizes are sampled dynamically according to an AR step decay strategy (details available in Appendix~\ref{sec:arstep}). For example, the model may select 2 tokens in the first AR step, another 2 in the second, and 3 in the third, as illustrated in Appendix Figure~\ref{fig:mask}. Further details are available in Appendix~\ref{sec:dep_mask}. This adaptive selection enables the model to effectively handle the non-uniform noise and sparsity patterns often present in microbiome profiles.

The integration of AR and diffusion is formally defined in Equation~(\ref{eq:cd-factorization}), where $\kappa_s$ denotes the subset of latent tokens sampled at the $s$-th AR step and $1 \leq s \leq S$ with $S$ as the total number of AR steps:
\begin{align}
q(\hat{x}_{mb}^{i,0:T,\kappa_s} \mid \hat{x}_{mb}^{i,0,\kappa_{1:s-1}}) 
&= q(\hat{x}_{mb}^{i,0,\kappa_s}) \notag \\
&\quad \times \prod_{t=1}^T q(\hat{x}_{mb}^{i,t,\kappa_s} \mid \hat{x}_{mb}^{i,t-1,\kappa_s}, \hat{x}_{mb}^{i,0,\kappa_{1:s-1}})
\label{eq:cd-factorization}
\end{align}

As depicted in Figure~\ref{fig:cat_diff}, each AR step $s$ processes a different subset $\kappa_s$ of latent tokens. During training, the model learns to approximate the reverse process $p_\theta(\hat{x}_{mb}^{i,t-1,\kappa_s} | \hat{x}_{mb}^{i,t,\kappa_s},\hat{x}_{mb}^{i,0,\kappa_{1:s-1}})$ for each diffusion step $t$ and AR step $s$, leveraging both the current noisy tokens and previously denoised ones. This hybrid framework effectively captures feature-level dependencies across multiple regulatory and microbe scales.
\paragraph{\bf Dependency-Aware Attention Mask}
To faithfully model microbial dependencies, we design a dependency-aware attention mask that integrates both directional dependencies and statistical associations among microbes. The dependency aware attention mask enforces correct sequential dependencies by restricting access such that $\hat{x}_{mb}^{i,t,\kappa_s}$ cannot be influenced by $\hat{x}_{mb}^{i,0,\kappa_{1:s-1}}$. This mechanism ensures that each autoregressive (AR) step relies only on clean tokens from prior steps, thereby preserving the dependency-aware ordering of microbial interactions. 

During training, clean tokens are appended after condition tokens, where only the first $S{-}1$ AR steps use the clean tokens for guidance. As shown in Figure~\ref{fig:method}, a typical configuration may include a condition token length of 2 and AR split sizes of [2, 2, 3]. The corresponding mask is constructed such that each AR segment is only allowed to attend to earlier segments and condition tokens, respecting both sequential and dependency constraints. Time-embedded noisy tokens are appended at the end and are processed by the Transformer module under this masking scheme.

To encode biological structure, we combine two dependency estimation techniques. First, we identify causal dependencies where feature $j$ influences feature $i$ by checking if including $x_j$ improves prediction of $y_i$. This is done by fitting a linear model $y_i \sim x_i + x_j$ and comparing it, via an F-statistic, against the restricted model $y_i \sim x_i$. A binary adjacency matrix $C_{\text{dir}} \in \{0,1\}^{D \times D}$ is then obtained by thresholding these F-statistics. Second, we estimate pairwise mutual information to uncover symmetric statistical dependencies, generating another binary mask $C_{\text{mi}} \in \{0,1\}^{D \times D}$ by applying a chosen threshold. The final dependency matrix $\mathbf{Dep} \in \{0,1\}^{D \times D}$ used in the attention mask is obtained by combining both structures:
\[\mathbf{Dep} = C_{\text{dir}} \lor C_{\text{mi}}\]
This hybrid matrix is then integrated into the attention masking process (Algorithm~\ref{alg:attn_mask}), guiding the Transformer to attend selectively to microbially relevant context across time and feature space.

The resulting attention mask $\mathbf{M}$ ensures that the diffusion model attends only to valid conditioning and context tokens while obeying both autoregressive temporal flow and microbial dependency structure. This enables the model to capture generalization across diverse microbiome datasets better, helping the model prioritize biologically plausible regulatory dependencies.

\paragraph{\bf Inference from Noisy Input}
During inference, the model denoises corrupted microbiome data by conditioning on both the partially observed microbial latent representations and encoded patient metadata. The process begins with the noisy latent representation $\hat{x}_{mb}^{i,T}$ obtained at the final diffusion step. Simultaneously, the observed latent portion $m_0^{c}$ extracted using a binary mask from the pretrained VAE encoder is fused with projected embeddings from patient metadata, which are encoded using a pretrained large language model. This combined conditioning signal guides the DAT module in estimating the noise component and predicting the denoised latent representation $\tilde{x}_{mb}^{i,T-1} \in \mathbb{R}^p$. At each reverse step $t = T, T{-}1, \dots, 1$, the DAT module integrates both causal structure and the conditioning information to progressively refine the microbial representation. The final output corresponds to the reconstructed microbiome data, completing the imputation process.

\paragraph{\bf Reverse Diffusion Process}
The reverse diffusion process aims to recover the clean latent microbiome representation by progressively removing noise from the corrupted input $\hat{x}_{mb}^{i,T}$. Starting from this noisy latent variable, the model applies a learned denoising function parameterized by $\theta$ (the parameters of our DAT module) to estimate $\hat{x}_{mb}^{i,t-1}$ from $\hat{x}_{mb}^{i,t}$ in an iterative manner. Each sampling step is governed by $p_\theta(\hat{x}_{mb}^{i,t-1} | \hat{x}_{mb}^{i,t}, c)$ for $t = T, T{-}1, \dots, 1$, where $\hat{x}_{mb}^{i,T}$ is initially sampled from a standard Gaussian distribution $\mathcal{N}(0, I)$ and $c$ denotes the conditioning information. Throughout this process, the conditioning vector $c$, comprising both the observed latent features $m_0^{c}$ and the pre-trained LLM-encoded metadata, remains fixed and is utilized to guide the generation toward biologically plausible reconstructions. The final result, $\hat{x}_{mb}^{i,0}$, reflects the denoised and imputed microbiome profile.
This section outlines the dataset and training configurations, followed by a comparative evaluation of the proposed DepMicroDiff approach.
\subsection{Dataset and Pre-processing}

All microbiome datasets used in this study were sourced from The Cancer Genome Atlas (TCGA) public repository~\cite{weinstein2013cancer, dohlman2021cancer}, covering various cancer types. For model training and evaluation, we selected datasets from three cancer types with relatively large sample sizes: Stomach Adenocarcinoma (STAD), Colon Adenocarcinoma (COAD), and Head and Neck Squamous Cell Carcinoma (HNSC). Additionally, datasets from Esophageal Carcinoma (ESCA) and Rectum Adenocarcinoma (READ), which have comparatively smaller sample sizes, were used solely for the pretraining phase. A detailed overview of these datasets is presented in Table~\ref{data}.
\begin{table}[htp]
\caption{The list of microbial datasets used in the study includes the three cancer types with the highest number of samples from TCGA.}
\label{data}
\begin{adjustbox}{width=.492\textwidth}
\fontsize{12}{12}\selectfont   
\setlength{\arrayrulewidth}{0.05mm} 
\renewcommand{\arraystretch}{1.5}
\begin{tabular}{l|cc|cc|c}
\hline
\multicolumn{1}{c|}{\multirow{2}{*}{Datasets}} & \multicolumn{2}{c|}{No. of Samples/Microbes}        & \multicolumn{2}{c|}{Prepro. Samples/Microbes}         & \multirow{2}{*}{Dropout Rate} \\ \cline{2-5}
\multicolumn{1}{c|}{}                          & \multicolumn{1}{c|}{\#Samples} & \#Microbes            & \multicolumn{1}{c|}{\#Samples} & \#Microbes           &                                \\ \hline
STAD                                           & \multicolumn{1}{c|}{530}       & \multirow{3}{*}{1289} & \multicolumn{1}{c|}{530}       & \multirow{3}{*}{106} & 87.61\%                        \\
COAD                                           & \multicolumn{1}{c|}{561}       &                       & \multicolumn{1}{c|}{561}       &                      & 63.20\%                        \\
HNSC                                           & \multicolumn{1}{c|}{587}       &                       & \multicolumn{1}{c|}{587}       &                      & 79.63\%                        \\ \hline
READ                                           & \multicolumn{1}{c|}{182}       & \multirow{2}{*}{1289} & \multicolumn{1}{c|}{182}       & \multirow{2}{*}{106} & 67.06\%                        \\
ESCA                                           & \multicolumn{1}{c|}{248}       &                       & \multicolumn{1}{c|}{248}       &                      & 88.99\%                        \\ \hline
\end{tabular}
\end{adjustbox}
\end{table}
Normalization of microbial abundance data is a critical yet challenging step in microbiome analysis. To ensure consistency and comparability across cancer types, we adopted a normalization strategy similar to that used in \cite{shi2024pretrained}. Specifically, for each individual cancer dataset, we performed row-wise normalization on the abundance matrix $M \in \mathbb{R}^{n \times d}$, where $n$ is the number of samples and $d$ is the number of microbial features. The normalized matrix is computed as:
\begin{equation}
M_{ij}^{'} = 10^{2} \cdot \frac{M_{ij}}{\sum_{k=1}^{d} M_{ik}},
\end{equation}
resulting in a transformed matrix $M' \in \mathbb{R}^{n \times d}$. To further reduce the influence of outlier values and enhance numerical stability, we applied a logarithmic transformation to $M'$, yielding the final input matrix $Y \in \mathbb{R}^{n \times d}$ defined as:
\begin{equation}
Y = \log_{10}(M_{ij}^{'} + 1.0).
\end{equation}
This two-step normalization and transformation process ensures that extreme variations in microbial counts are attenuated, facilitating robust learning by our model.

\subsection{Baseline Models}
To establish a robust comparison, our work benchmarks against eight representative methods: KNN, DeepImpute \cite{deepimpute}, AutoImpute \cite{autoimpute}, DCA \cite{dca}, CpG Transformer \cite{cpg}, scVI \cite{scvi}, DeepMicroGen \cite{deepmicrogen}, and mbVDiT \cite{shi2024pretrained}. These methods collectively offer diverse assumptions and inductive biases, allowing us to evaluate how well each captures the complex and sparse distributions characteristic of microbiome data.
\begin{figure}[H]
    \centering
    \includegraphics[width=0.35\textwidth]{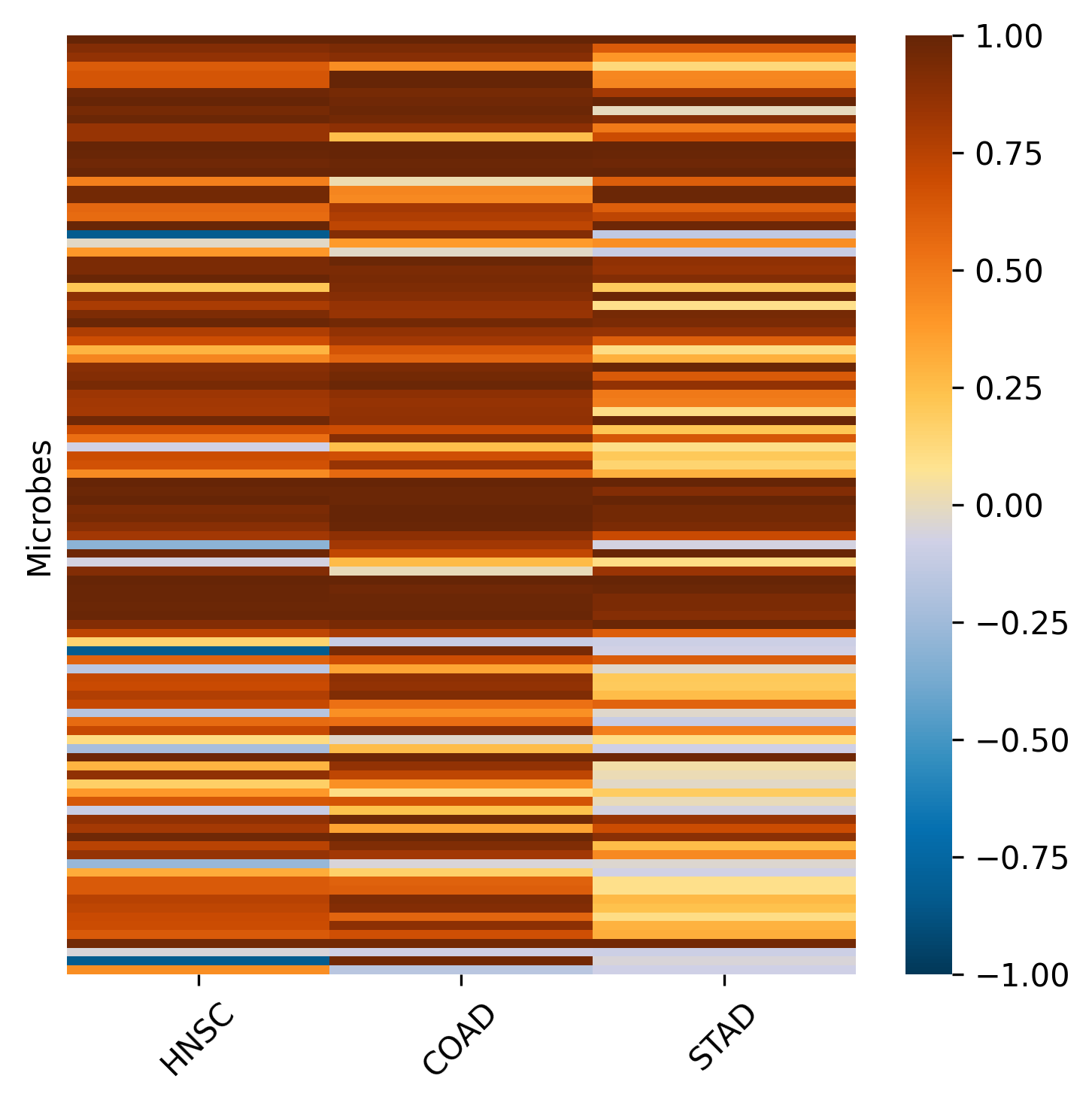}
    \caption{Heatmap of Pearson correlation coefficients between imputed and real microbiome data.}
    \label{fig:heatmap}
\end{figure}
\subsection{Implementation Details}
DepMicroDiff is implemented using the PyTorch framework~\cite{paszke2019pytorch} and is trained on NVIDIA A100 GPUs with optimized hyperparameters. The model architecture is adapted from the DAT framework, redesigned to address the unique characteristics of microbiome data. It employs a dependency-aware attention mechanism specifically tailored for microbial features, enabling effective modeling of regulatory dependencies. The input data is tokenized such that each token corresponds to a microbe-level compressed representation. To enhance generalization across datasets, we apply a VAE-based pretraining scheme using microbiome profiles from multiple cancer types. During training and inference, the model conditions on both the partially observed latent features and LLM-encoded patient metadata (e.g., sample type, stage), guiding the denoising process with biologically relevant context.
\subsection{Evaluation Metrics}
To assess the effectiveness of DepMicroDiff and the baseline models, we employed four evaluation metrics Pearson Correlation Coefficient (PCC), Cosine Similarity (COS), Root Mean Square Error (RMSE), and Mean Absolute Error (MAE) across three different datasets. 

\subsection{Experimental Results}
The detailed experimental result is showing in Table~\ref {tab2}. To demonstrate the effectiveness of our model, we computed the Pearson correlation coefficients (PCCs) between the imputed and real microbiome data. Specifically, for each dataset (HNSC, COAD, STAD), we calculated PCCs for individual microbes and visualized the resulting correlation values as a heatmap. In Table~\ref{tab2}, the structured and consistently high correlations across all three datasets illustrate that DepMicroDiff reliably reconstructs biologically meaningful microbial signals, capturing both global patterns and dataset-specific variations. Figure ~\ref{fig:heatmap} visualizes these correlations as a heatmap, where each row corresponds to a microbe and each column represents a cancer type. Warmer colors (toward 1.0) indicate strong positive correlations, while cooler tones reflect lower or even negative correlations. As shown in the figure, the majority of microbes across all datasets exhibit high PCC values, suggesting that DepMicroDiff successfully reconstructs biologically meaningful microbial structures. The consistent intensity patterns across datasets also indicate the model’s robustness and generalizability in handling different microbiome profiles.  

\begin{figure}[t]
    \centering
    \includegraphics[width=0.5\textwidth]{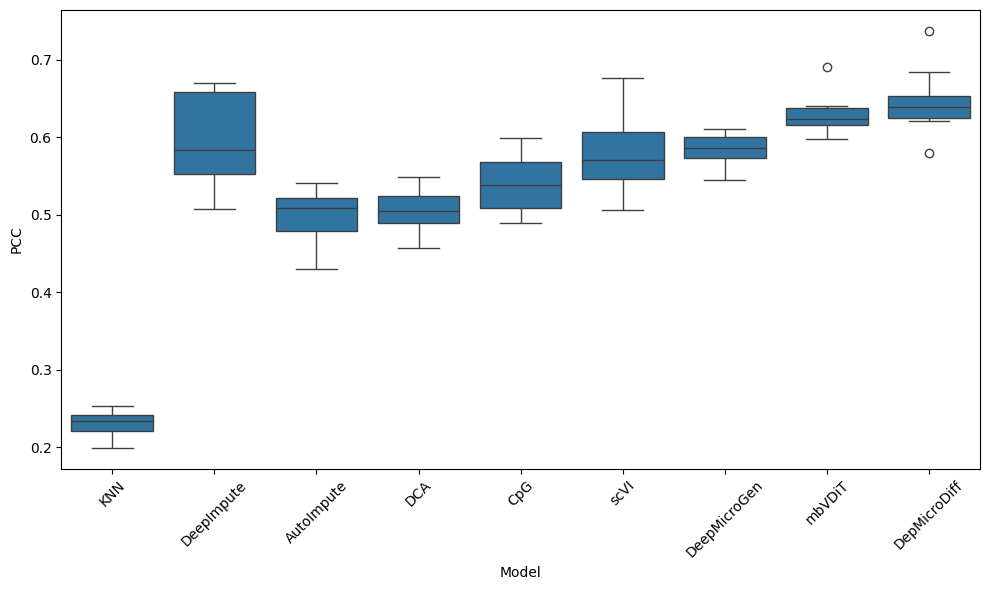}
    \caption{Boxplot of Pearson correlation coefficients between imputed microbiome
 data and real microbiome data.}
    \label{fig:pcc}
\vspace{-10pt}
\end{figure}

Figure~\ref{fig:pcc} presents a boxplot comparison of Pearson correlation coefficients (PCCs) between imputed and ground truth microbiome data across various imputation methods. Each box reflects the distribution of PCC values across different microbes, highlighting both the central tendency and variance of each model's performance. Our proposed method, DepMicroDiff, consistently achieves the highest median PCC and exhibits a tighter distribution compared to all baselines, including mbVDiT\cite{shi2024pretrained}, DeepMicroGen\cite{deepmicrogen}, and scVI\cite{scvi}. Furthermore, DepMicroDiff shows lower RMSE and MAE, demonstrating its robustness and generalizability for microbiome imputation. This result underscores the effectiveness of integrating dependency modeling and patient metadata in enhancing the fidelity of microbiome data reconstruction.

\begin{table*}[htp]
\centering
\caption{Performance comparison between DepMicroDiff and baselines on three microbiome datasets (STAD, COAD, and HNSC) using four evaluation metrics. The best performance for each metric is shown in bold, and the second-best is underlined.}
\label{tab2}
\begin{adjustbox}{width=1\textwidth}
\fontsize{9}{11}\selectfont         
% \rowcolors{2}{gray!10}{white}  % Alternate row color
\begin{tabular}{l|ccccccccc}
\toprule
\textbf{PCC $\uparrow$}  & KNN & DeepImpute & AutoImpute & DCA & CpG & scVI & DeepMicroGen & mbVDiT & DepMicroDiff \\ 
\midrule
\textbf{STAD} & 0.232±0.021 & 0.570±0.067 & 0.477±0.048 & 0.505±0.025 & 0.520±0.049 & 0.596±0.065 & 0.588±0.065 & 0.634±0.032 & \textbf{0.641±0.046} \\
\textbf{COAD} & 0.188±0.033 & 0.653±0.054 & 0.639±0.085 & 0.632±0.062 & 0.599±0.069 & 0.667±0.059 & 0.675±0.049 & 0.704±0.062 & \textbf{0.712±0.011} \\
\textbf{HNSC} & 0.247±0.038 & 0.592±0.032 & 0.550±0.043 & 0.594±0.056 & 0.585±0.017 & 0.559±0.028 & 0.604±0.061 & 0.626±0.060 & \textbf{0.636±0.019} \\
\midrule
\textbf{Cosine $\uparrow$} & KNN & DeepImpute & AutoImpute & DCA & CpG & scVI & DeepMicroGen & mbVDiT & DepMicroDiff \\
\midrule
\textbf{STAD} & 0.430±0.014 & 0.773±0.059 & 0.787±0.027 & 0.746±0.036 & 0.746±0.027 & 0.775±0.072 & 0.775±0.074 & 0.806±0.057 & \textbf{0.812±0.039} \\
\textbf{COAD} & 0.241±0.037 & 0.769±0.057 & 0.650±0.124 & 0.724±0.064 & 0.688±0.062 & 0.716±0.052 & 0.772±0.072 & \textbf{0.791±0.080} & \underline{0.789±0.017} \\
\textbf{HNSC} & 0.357±0.031 & 0.779±0.044 & 0.794±0.050 & 0.789±0.026 & 0.778±0.015 & 0.776±0.024 & 0.786±0.031 & \textbf{0.802±0.062} & \underline{0.798±0.018} \\
\midrule
\textbf{RMSE $\downarrow$} & KNN & DeepImpute & AutoImpute & DCA & CpG & scVI & DeepMicroGen & mbVDiT & DepMicroDiff \\
\midrule
\textbf{STAD} & 3.371±0.124 & 1.572±0.082 & 1.521±0.055 & 1.629±0.044 & 1.462±0.075 & 2.478±0.085 & 1.469±0.049 & 1.320±0.053 & \textbf{1.290±0.022} \\
\textbf{COAD} & 4.336±0.106 & 1.211±0.076 & 1.370±0.070 & 1.183±0.059 & 1.127±0.045 & 2.351±0.057 & 1.002±0.100 & 0.934±0.054 & \textbf{0.927±0.064} \\
\textbf{HNSC} & 4.128±0.083 & 1.258±0.058 & 1.364±0.049 & 1.246±0.034 & 1.169±0.022 & 2.168±0.077 & 1.245±0.047 & 1.155±0.027 & \textbf{1.098±0.013} \\
\midrule
\textbf{MAE $\downarrow$} & KNN & DeepImpute & AutoImpute & DCA & CpG & scVI & DeepMicroGen & mbVDiT & DepMicroDiff \\
\midrule
\textbf{STAD} & 3.255±0.095 & 1.388±0.061 & 1.186±0.039 & 1.234±0.047 & 1.308±0.071 & 2.221±0.068 & 1.325±0.058 & 0.956±0.050 & \textbf{0.933±0.086} \\
\textbf{COAD} & 3.862±0.114 & 0.804±0.065 & 0.864±0.128 & 0.728±0.061 & 0.739±0.063 & 2.132±0.049 & 0.627±0.109 & 0.530±0.070 & \textbf{0.524±0.012} \\
\textbf{HNSC} & 3.679±0.096 & 0.986±0.064 & 0.981±0.045 & 0.993±0.018 & 0.906±0.021 & 0.847±0.083 & 0.978±0.028 & 0.807±0.020 & \textbf{0.799±0.021} \\
\bottomrule
\end{tabular}
\end{adjustbox}
\end{table*}
\section{Ablation Study}
In this section, we remove some components of our model to assess each component’s effectiveness towards the whole model, with the evaluation metrics.
\raggedbottom
\subsection{VAE Pretraining}
To overcome the limitations posed by the relatively small sample sizes within individual cancer-type microbiome datasets, we adopt a VAE-based pretraining strategy to enhance model performance. Instead of training solely on a single dataset, which often results in suboptimal generalization, we leverage microbiome data from other cancer types to learn a robust, shared weight initialization. These pretrained parameters are then transferred to the VAE module in \textit{DepMicroDiff}, providing a strong initialization for downstream fine-tuning.

\begin{figure}[H]
    \centering
    \includegraphics[width=0.5\textwidth]{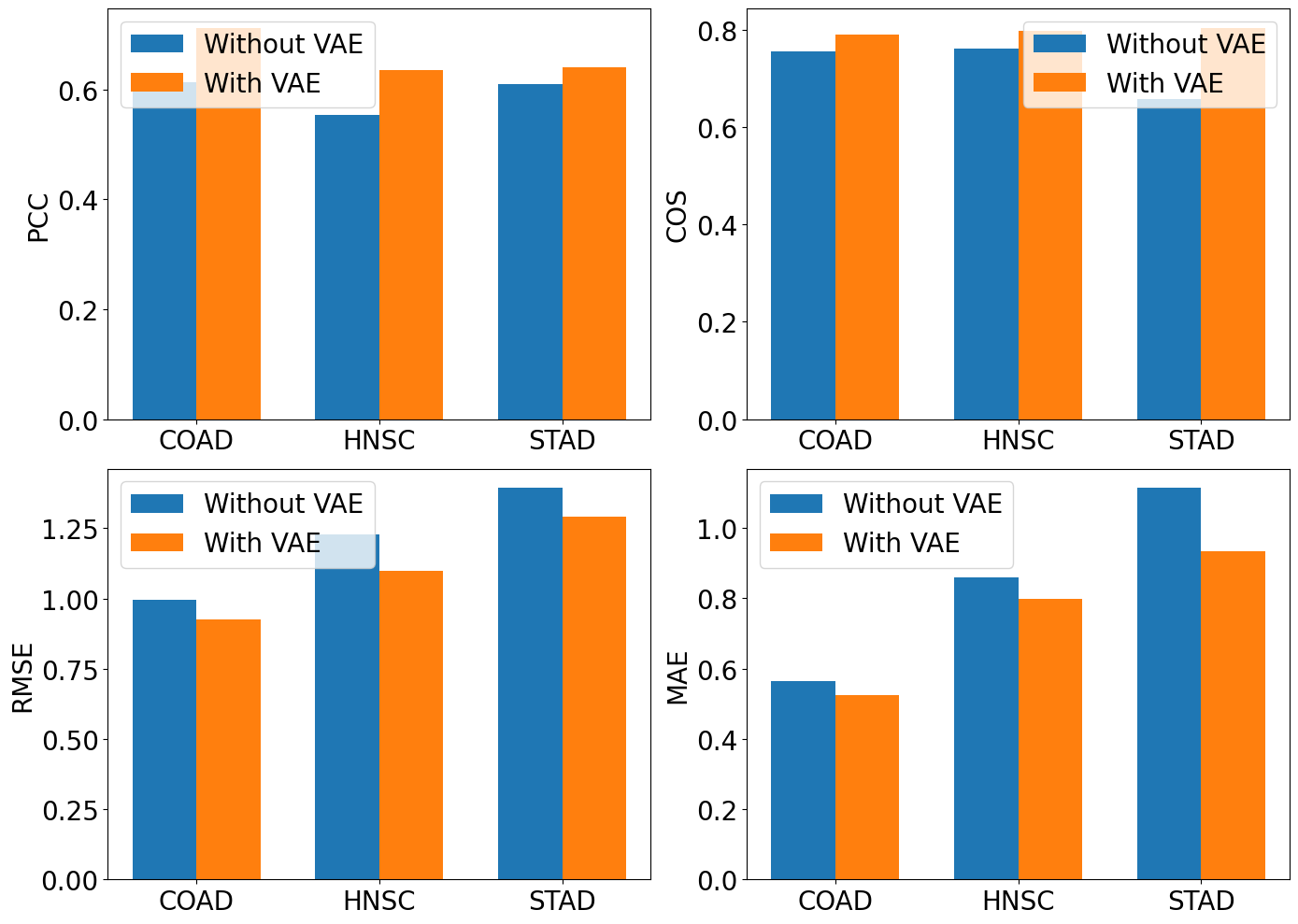}
    \caption{
        Performance comparison between models trained with and without VAE pretraining across three microbiome datasets (COAD, HNSC, STAD) using four evaluation metrics: PCC, COS, RMSE, and MAE. VAE pretraining consistently improves both correlation-based and error-based metrics, indicating enhanced reconstruction capability.
    }
    \label{fig:vae-pretrain}
\end{figure}

As shown in Figure~\ref{fig:vae-pretrain}, comparative results across three datasets (COAD, HNSC, and STAD) consistently demonstrate that VAE pretraining yields noticeable improvements across all evaluation metrics, including PCC, COS, RMSE, and MAE. These findings confirm that incorporating pretraining not only enhances the imputation quality but also mitigates the challenges associated with limited data availability in individual datasets. This validates the effectiveness of our transfer learning approach for learning generalized microbial representations across cancer types.

\raggedbottom
\subsection{Inclusion of Metadata}
\begin{figure}[H]
    \centering
    \includegraphics[width=0.75\linewidth]{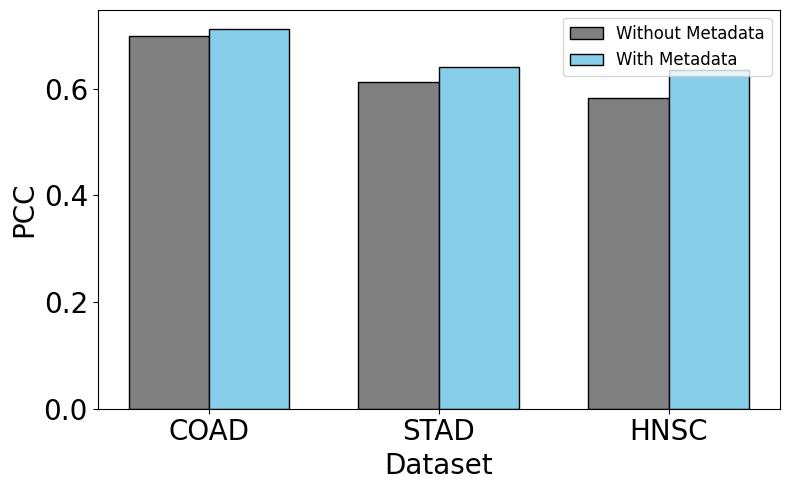} % Update with your actual path
    \caption{Comparison of PCC values on COAD, STAD, and HNSC datasets with and without metadata conditioning. Incorporating patient-level metadata consistently improves imputation performance.}
    \label{fig:metadata_ablation}
\end{figure}

To assess the contribution of patient metadata in our model, we performed ablation experiments by comparing versions of the model with and without metadata conditioning. As shown in Figure~\ref{fig:metadata_ablation}, incorporating patient-specific information such as sample type, pathologic stage, and age into the diffusion model leads to consistent improvements in Pearson correlation coefficient (PCC) across all three cancer datasets. These gains indicate that metadata provides informative context that helps the model learn biologically meaningful imputations. From the result, we can validate the utility of leveraging auxiliary clinical information to enhance model performance.

\section{Conclusion}
We present DepMicroDiff, a novel diffusion-based framework that addresses critical challenges in microbiome data imputation. By integrating a Dependency-Aware Transformer with diffusion modeling, our approach captures complex interdependencies among microbial taxa that existing methods overlook. VAE-based pretraining enables cross-tissue generalization, while LLM-encoded metadata provides patient-specific context for improved imputation accuracy.
Extensive experiments on TCGA datasets show that DepMicroDiff substantially outperforms state-of-the-art baselines, validating our framework's effectiveness in handling the extreme sparsity and structured dependencies characteristic of microbiome data. By enabling more accurate microbiome imputation, our work facilitates critical downstream analyses for understanding host-microbiome interactions and advancing precision medicine. Future work could extend DepMicroDiff to temporal microbiome analysis, explore the interpretability of learned dependencies for ecological insights, and validate its applicability to other sparse biological data types.

% -------------------- Acknowledgments --------------------
\section*{Acknowledgment}
This research is supported in part by the NSF under Grant IIS 2327113 and ITE 2433190 and the NIH under Grants R21AG070909 and P30AG072946. We would like to thank NSF for support for the AI research resource with NAIRR240219. We thank the University of Kentucky Center for Computational Sciences and Information Technology Services Research Computing for their support and use of the Lipscomb Compute Cluster and associated research computing resources. We also acknowledge and thank those who created, cleaned, and curated the datasets used in this study.

% -------------------- References --------------------
\bibliographystyle{IEEEtran}
\bibliography{sample-base}

\begin{thebibliography}{10}
\providecommand{\url}[1]{#1}
\csname url@samestyle\endcsname
\providecommand{\newblock}{\relax}
\providecommand{\bibinfo}[2]{#2}
\providecommand{\BIBentrySTDinterwordspacing}{\spaceskip=0pt\relax}
\providecommand{\BIBentryALTinterwordstretchfactor}{4}
\providecommand{\BIBentryALTinterwordspacing}{\spaceskip=\fontdimen2\font plus
\BIBentryALTinterwordstretchfactor\fontdimen3\font minus \fontdimen4\font\relax}
\providecommand{\BIBforeignlanguage}[2]{{%
\expandafter\ifx\csname l@#1\endcsname\relax
\typeout{** WARNING: IEEEtran.bst: No hyphenation pattern has been}%
\typeout{** loaded for the language `#1'. Using the pattern for}%
\typeout{** the default language instead.}%
\else
\language=\csname l@#1\endcsname
\fi
#2}}
\providecommand{\BIBdecl}{\relax}
\BIBdecl

\bibitem{dohlman2021cancer}
A.~B. Dohlman, D.~A. Mendoza, S.~Ding, M.~Gao, H.~Dressman, I.~D. Iliev, S.~M. Lipkin, and X.~Shen, ``The cancer microbiome atlas: a pan-cancer comparative analysis to distinguish tissue-resident microbiota from contaminants,'' \emph{Cell host \& microbe}, vol.~29, no.~2, pp. 281--298, 2021.

\bibitem{deepimpute}
C.~Arisdakessian, O.~Poirion, B.~Yunits, X.~Zhu, and L.~X. Garmire, ``{DeepImpute: An Accurate, Fast, and Scalable Deep Neural Network Method to Impute Single-Cell RNA-seq Data},'' \emph{Genome Biology}, vol.~20, pp. 1--14, 2019.

\bibitem{dca}
G.~Eraslan, L.~M. Simon, M.~Mircea, N.~S. Mueller, and F.~J. Theis, ``{Single-Cell RNA-seq Denoising Using a Deep Count Autoencoder},'' \emph{Nature Communications}, vol.~10, no.~1, p. 390, 2019.

\bibitem{scvi}
R.~Lopez, J.~Regier, M.~B. Cole, M.~I. Jordan, and N.~Yosef, ``{Deep Generative Modeling for Single-Cell Transcriptomics},'' \emph{Nature Methods}, vol.~15, no.~12, pp. 1053--1058, 2018.

\bibitem{shi2024pretrained}
X.~Shi, F.~Zhu, and W.~Min, ``Pretrained-guided conditional diffusion models for microbiome data analysis,'' in \emph{2024 IEEE International Conference on Bioinformatics and Biomedicine (BIBM)}.\hskip 1em plus 0.5em minus 0.4em\relax IEEE, 2024, pp. 579--584.

\bibitem{zhong2024synthesizing}
Y.~Zhong, X.~Wang, J.~Wang, X.~Zhang, Y.~Wang, M.~Huai, C.~Xiao, and F.~Ma, ``Synthesizing multimodal electronic health records via predictive diffusion models,'' in \emph{Proceedings of the 30th ACM SIGKDD Conference on Knowledge Discovery and Data Mining}, 2024, pp. 4607--4618.

\bibitem{senane2024self}
Z.~Senane, L.~Cao, V.~L. Buchner, Y.~Tashiro, L.~You, P.~A. Herman, M.~Nordahl, R.~Tu, and V.~Von~Ehrenheim, ``Self-supervised learning of time series representation via diffusion process and imputation-interpolation-forecasting mask,'' in \emph{Proceedings of the 30th ACM SIGKDD Conference on Knowledge Discovery and Data Mining}, 2024, pp. 2560--2571.

\bibitem{autoimpute}
D.~Talwar, A.~Mongia, D.~Sengupta, and A.~Majumdar, ``{AutoImpute: Autoencoder Based Imputation of Single-Cell RNA-seq Data},'' \emph{Scientific Reports}, vol.~8, no.~1, p. 16329, 2018.

\bibitem{cpg}
G.~De~Waele, J.~Clauwaert, G.~Menschaert, and W.~Waegeman, ``{CpG Transformer for Imputation of Single-Cell Methylomes},'' \emph{Bioinformatics}, vol.~38, no.~3, pp. 597--603, 2022.

\bibitem{deepmicrogen}
J.~M. Choi, M.~Ji, L.~T. Watson, and L.~Zhang, ``{DeepMicroGen: A Generative Adversarial Network-Based Method for Longitudinal Microbiome Data Imputation},'' \emph{Bioinformatics}, vol.~39, no.~5, p. btad286, 2023.

\bibitem{bucci2016microbial}
V.~Bucci, B.~Tzen, N.~Li, S.~Simmons, T.~Tanoue, E.~Bogart, L.~Deng, V.~Yeliseyev, M.~Delaney, J.~Liu \emph{et~al.}, ``Mdsine: Microbial dynamical systems inference engine for microbiome time-series analyses,'' \emph{Genome Biology}, vol.~17, no.~1, p. 121, 2016.

\bibitem{deng2024causal}
C.~Deng, D.~Zh, K.~Li, S.~Guan, and H.~Fan, ``Causal diffusion transformers for generative modeling,'' \emph{arXiv preprint arXiv:2412.12095}, 2024.

\bibitem{weinstein2013cancer}
C.~J.~C. Kyle~Chang \emph{et~al.}, ``The cancer genome atlas pan-cancer analysis project,'' \emph{Nature Genetics}, vol.~45, no.~10, pp. 1113--1120, 2013.

\bibitem{paszke2019pytorch}
A.~Paszke, S.~Gross, F.~Massa, A.~Lerer, J.~Bradbury, G.~Chanan, T.~Killeen, Z.~Lin, N.~Gimelshein, L.~Antiga \emph{et~al.}, ``Pytorch: An imperative style, high-performance deep learning library,'' \emph{Advances in neural information processing systems}, vol.~32, 2019.

\end{thebibliography}
\raggedbottom
\section*{Appendix}
\subsection{Dependency-Aware Mask formation}
\label{sec:dep_mask}
\begin{figure}[H]
    \centering
    \includegraphics[width=0.75\linewidth]{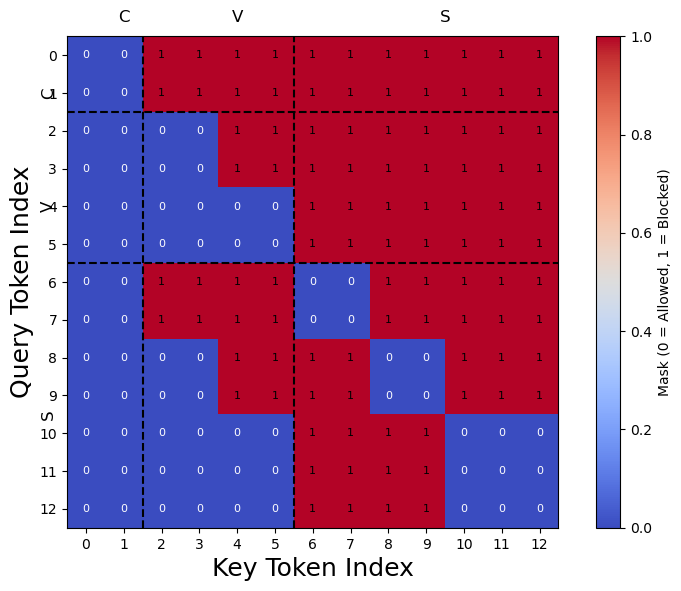} % Update with your actual path
    \caption{Blockwise visualization of the generalized dependency-aware attention mask. The attention matrix is divided into three semantic blocks: \textbf{C} (Conditional tokens), \textbf{V} (Visible tokens from previous autoregressive steps), and \textbf{S} (Current autoregressive tokens). A value of \texttt{0} (blue) indicates allowed attention, while \texttt{1} (red) indicates blocked connections. This design enables autoregressive denoising while preserving dependency across blocks.}
    \label{fig:mask}
\end{figure}

The construction of the dependency-guided attention mask begins with Algorithm~\ref{alg:split_integer_exp_decay}, which returns the split sizes $sz$ and their cumulative sums $cs$ as inputs to Algorithm~\ref{alg:attn_mask}.
A detailed illustration of this masking process is provided in Figure~\ref{fig:mask}. 

For instance, consider a sample length of $s = 7$ and a conditional length of $c = 2$, with split sizes $sz = [2,2,3]$ and cumulative sums $cs = [0,2,4,7]$. From these values, the key lengths are calculated as follows: visible length $v = 4$, context length $ctx = 6$, and total sequence length $seq = 13$. A $13 \times 13$ attention mask matrix is initialized with ones, and the first two columns are set to zero to make the condition tokens visible across the sequence. The remaining matrix is then partitioned into three logical blocks: visible-to-visible ($vTv$), sample-to-visible ($sTv$), and sample-to-sample ($sTs$), which are further refined using autoregressive rules and optionally a dependency mask.
\begin{algorithm}[htp]
\caption{Generate AR Steps}
\label{alg:split_integer_exp_decay}
\textbf{Input:} $S$ (sample length to split), $\alpha$ (exponential decay factor)
\vspace{-0.4cm}
\begin{algorithmic}[1]
\IF{$\alpha = 1.0$}
    \STATE $N \gets \text{random integer in } [1, S]$
\ELSE
    \STATE $b \gets \frac{1 - \alpha}{1 - \alpha^S}$ \hfill {\color{gray}\texttt{\% Normalization factor}}
    \STATE $p \gets [b \cdot \alpha^i \text{ for } i \in [0, S-1]]$
    \STATE $N \gets \text{random choice from } [1, S] \text{ with probabilities } p$
\ENDIF

\STATE \textbf{Generate cumulative sum:}
\STATE $cs \gets [0] + \text{sort(random sample from } [1, S) \text{ of size } N-1) + [S]$
\STATE $sz \gets [cs[i+1] - cs[i] \text{ for } i \in [0, N-1]]$

\STATE \textbf{return} $sz, cs$
\end{algorithmic}
\end{algorithm}

\begin{algorithm}[H]
\caption{Generate Dependency-Aware Attention Mask}
\label{alg:attn_mask}

\textbf{Input:} $s$ (sample length), $c$ (conditional length), $sz$ (split sizes), $cs$ (cumulative sum of split sizes), $\mathbf{Dep} \in \{0,1\}^{D \times D}$ (dependency mask)

\begin{algorithmic}[1]
\STATE $v \gets s - sz[-1]$ \hfill {\color{gray}\texttt{\% Visible length}}
\STATE $ctx \gets c + v$ \hfill {\color{gray}\texttt{\% Context length}}
\STATE $seq \gets ctx + s$ \hfill {\color{gray}\texttt{\% Total sequence length}}

\STATE Initialize $\mathbf{M} \in \mathbb{R}^{seq \times seq}$ with ones
\STATE $\mathbf{M}[:,:c] \gets 0$ \hfill {\color{gray}\texttt{\% Condition tokens visible to all}}

\STATE \textbf{Initialize block masks:}
\STATE $vTv \gets \mathbf{1}^{v \times v}$ \hfill {\color{gray}\texttt{\% Visible to visible}}
\STATE $sTv \gets \mathbf{1}^{s \times v}$ \hfill {\color{gray}\texttt{\% Sample to visible}}
\STATE $sTs \gets \mathbf{1}^{s \times s}$ \hfill {\color{gray}\texttt{\% Sample to sample}}

\FOR{$i = 0$ \TO $|sz| - 2$}
    \STATE $vTv[cs[i]:cs[i+1], 0:cs[i+1]] \gets 0$
    \STATE $sTv[cs[i+1]:cs[i+2], 0:cs[i+1]] \gets 0$
\ENDFOR

\FOR{$i = 0$ \TO $|sz| - 1$}
    \STATE $sTs[cs[i]:cs[i+1], cs[i]:cs[i+1]] \gets 0$
\ENDFOR

\STATE \textbf{Incorporate dependency mask:}
\STATE $sTs \gets sTs \lor \mathbf{Dep}$ \hfill {\color{gray}\texttt{\% Combine with dependency mask}}

\STATE \textbf{Assign final attention mask:}
\STATE $\mathbf{M}[c:ctx, c:ctx] \gets vTv$
\STATE $\mathbf{M}[ctx:, c:ctx] \gets sTv$
\STATE $\mathbf{M}[ctx:, ctx:] \gets sTs$

\STATE \textbf{return} $\mathbf{M}$
\end{algorithmic}
\end{algorithm}

\subsection{AR Step Decay}  
\label{sec:arstep}
In this section, we analyze the impact of the AR-step decay rate on the performance of our DepMicroDiff model. The AR-step decay controls the relative influence of earlier versus later autoregressive (AR) steps during denoising, mimicking the biological principle that upstream regulators often have a stronger impact on downstream elements in microbial regulatory pathways.

\begin{figure}[H]
    \centering
    \includegraphics[width=0.8\linewidth]{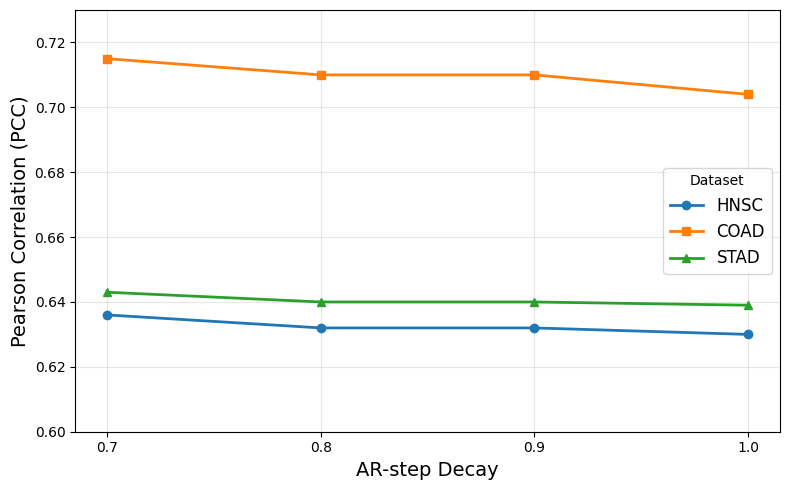}
    \caption{Effect of AR-step decay on model performance. The plot shows the Pearson correlation coefficients (PCCs) across different decay rates for three datasets (HNSC, COAD, and STAD).}
    \label{fig:arstep_decay}
\end{figure}

As shown in Figure~\ref{fig:arstep_decay}, we evaluated model performance in terms of Pearson correlation coefficient (PCC) across three datasets HNSC, COAD, and STAD under varying decay rates from 0.7 to 1.0. We observe a slight but consistent decline in PCC as the decay parameter approaches 1.0, where all AR steps are treated with equal importance. This trend indicates that our biologically inspired decay schedule, which emphasizes earlier AR steps, enhances the model's ability to reconstruct missing microbiome data.
Notably, the COAD dataset shows the highest PCC values and the most noticeable decline as decay increases, suggesting that datasets with more structured or hierarchical microbial relationships benefit more from stepwise decaying influence. On the other hand, HNSC and STAD exhibit relatively stable but slightly decreasing trends, indicating a more modest dependency on temporal step weighting. These results support the effectiveness of incorporating decaying influence into the AR framework, with a decay rate around 0.7–0.8 offering a good balance between step informativeness and predictive performance across varying microbial datasets.
\end{document}